\documentclass{ecai} 

\usepackage{latexsym}
\usepackage{amssymb}
\usepackage{amsmath}
\usepackage{amsthm}
\usepackage{booktabs}
\usepackage{enumitem}
\usepackage{graphicx}
\usepackage{color}
\usepackage{inconsolata}
\usepackage{amsmath}
\usepackage{booktabs}
\usepackage{graphicx}
\usepackage{pifont}
\usepackage{algorithm}
\usepackage{algorithmicx}
\usepackage{algpseudocode}
\usepackage{tabulary}

\usepackage{multirow}
\usepackage{makecell}
\usepackage{textcomp}

\usepackage{caption}
\newcommand{\ignore}[1]{{}}

\newcommand{\BibTeX}{B\kern-.05em{\sc i\kern-.025em b}\kern-.08em\TeX}

\begin{document}


\begin{frontmatter}


\paperid{2374} 


\title{Diversifying the Mixture-of-Experts Representation for Language Models with Orthogonal Optimizer}

\author{
        Boan~Liu, Liang~Ding, Li~Shen, Keqin~Peng, Yu~Cao, Dazhao~Cheng, Dacheng~Tao\thanks{B. Liu is with Hong Kong Polytechnic University, Hong Kong, China (e-mail:bo-an.liu@connect.polu.hk); D. Cheng is with School of Computer Science, Wuhan University, Wuhan, China (e-mail: dcheng@whu.edu.cn); L. Ding is with The University of Sydney, Australia (e-mail: liangding.liam@gmail.com); L. Shen is with the Sun Yat-sen University, Shenzhen, China (e-mail: mathshenli@gmail.com); Y. Cao is with Tencent IEG, Shenzhen, China (e-mail: rainyucao@tencent.com); D. Tao is with the Nanyang Technological University, Singapore (e-mail: dacheng.tao@gmail.com). Liang Ding and Dazhao Cheng are corresponding authors.}
}




\begin{abstract}
The Mixture of Experts (MoE) has emerged as a highly successful technique in deep learning, based on the principle of divide-and-conquer to maximize model capacity without significant additional computational cost. 
Even in the era of large-scale language models (LLMs), MoE continues to play a crucial role, as some researchers have indicated that GPT-4 adopts the MoE structure to ensure diverse inference results.
However, MoE is susceptible to performance degeneracy, particularly evident in the issues of imbalance and homogeneous representation among experts. While previous studies have extensively addressed the problem of imbalance, the challenge of homogeneous representation remains unresolved.
In this study, we shed light on the homogeneous representation problem, wherein experts in the MoE fail to specialize and lack diversity, leading to frustratingly high similarities in their representations (up to 99\% in a well-performed MoE model). This problem restricts the expressive power of the MoE and, we argue, contradicts its original intention.
To tackle this issue, we propose a straightforward yet highly effective solution: OMoE, an orthogonal expert optimizer. Additionally, we introduce an alternating training strategy that encourages each expert to update in a direction orthogonal to the subspace spanned by other experts.
Our algorithm facilitates MoE training in two key ways: firstly, it explicitly enhances representation diversity, and secondly, it implicitly fosters interaction between experts during orthogonal weights computation.
Through extensive experiments, we demonstrate that our proposed optimization algorithm significantly improves the performance of fine-tuning the MoE model on the GLUE benchmark, SuperGLUE benchmark, question-answering task, and name entity recognition tasks.
\end{abstract}

\end{frontmatter}


\section{Introduction}

The Mixture of Experts (MoE)~\cite{jacobs1991adaptive} is a widely adopted machine learning technique that puts several \textit{experts} in one model.
Essentially, each expert focuses on a particular field, and a gating network brings them together. 
This network picks the most appropriate expert for any given input, which has greatly enhanced the performance compared with single and general models.
Consequently, the MoE has been used in various domains, including machine translation~\cite{ding-tao-2019-university,ding2020understanding,ding2022redistributing}, sentiment analysis~\cite{wang2022contrastive}, dialogue~\cite{wang-etal-2023-divide}, and natural language generation~\cite{chai2023improved}.

\begin{figure}[h]
\centering
\includegraphics[width=0.92\linewidth]{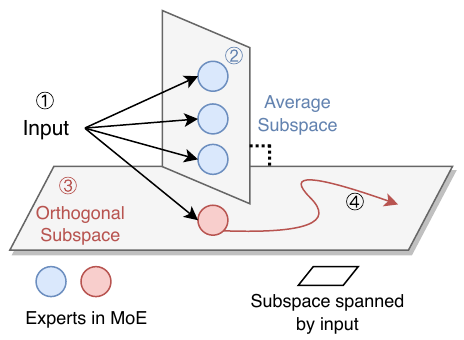}
\caption{\textbf{The overview of OMoE optimizer.} \ding{172} After being selected by the Gating Function, the input is sent to different experts. \ding{173} Experts calculate the corresponding orthogonal projector based on their input. \ding{174} Based on the orthogonal projectors of the other experts (\textit{e.g.} blue expert), the current expert to be updated (\textit{e.g.} red expert) calculates the average projector. The average projector represents the orthogonal subspace of other experts. \ding{175} Using the projector calculated in the previous step, the parameters are updated in the orthogonal direction of the other experts.}
\label{fig:overview}
\end{figure}

While the MoE has shown promising results in improving machine learning models' accuracy, it is prone to have performance degeneracy~\cite{shen2019mixture}, a phenomenon where the MoE model fails to outperform a single model with fewer parameters. There are two forms of degeneracy. The first form arises when only one expert dominates, resulting in the \textit{imbalance problem} where the rich get richer.
To address this issue, gating functions~\cite{roller2021hash,DBLP:conf/iclr/Zuo00KHZGZ22,fedus2021switch} and regularization techniques such as weight decay on loss functions~\cite{fedus2021switch} have been proposed.
The second form of degeneracy occurs when the experts fail to specialize and lack diversity, leading to the \textit{homogeneous representation problem}. 
Researchers~\cite{DBLP:conf/coling/Gao0ZLW22} propose a parameter-efficient MoE architecture that shares partial parameters among all experts to reduce memory consumption while providing auxiliary parameters for each expert to maintain diversity. Although several studies have addressed the issue of imbalanced expert load in MoE models, the second type of degeneracy caused by the homogeneous representation problem remains unresolved~\cite{DBLP:conf/iclr/Zuo00KHZGZ22,yang2021m6}.

To tackle the limited diversity problem among experts in MoE models, we propose a novel optimizer, the \textbf{OMoE Optimizer}.
The OMoE optimizer aims to increase expert diversity and improve performance. Our approach divides the training process into two alternating phases, each employing a specific optimizer.
In the first accumulating phase, a standard optimizer such as SGD, Adam~\cite{DBLP:journals/corr/KingmaB14}, or AdamW~\cite{DBLP:conf/iclr/LoshchilovH19} updates the model parameters, excluding the experts in MoE. Meanwhile, the OMoE optimizer collects input data for later use. 
In the second orthogonal phase, the OMoE optimizer updates the orthogonal projector based on the accumulated inputs, as depicted in Figure~\ref{fig:overview}.
To evaluate our method, we conducted comprehensive experiments, including fine-tuning pre-trained language models on the GLUE benchmark, SuperGLUE benchmark, Question-Answering (QA) task, and Name Entity Recognition (NER) task. Our \textbf{contributions} are summarized as follows:
\begin{itemize}
    \item We identify the crucial issue of degeneracy, which arises from the homogeneous representation problem in MoE. Both forms of degeneracy—where either one expert dominates or all experts become similar—severely degrade the performance of MoE models.
    
    \item Inspired by recent advancements in continual learning, we propose the OMoE Optimizer along with an alternative training approach to effectively enlarge the diversity among experts in MoE, which establishes a novel connection between tasks in continual learning and mini-batches in MoE training.  
    
    \item We evaluate OMoE upon several language models in fine-tuning GLUE benchmark, SuperGLUE benchmark, QA and NER task. 
    
\end{itemize}

\section{Background}
\subsection{Mixture of Expert}
An MoE model contains a set of expert network $E_1,E_2,......$ and a gating function $G$. Each expert specializes in different inputs that are decided by the gating function:

\begin{equation}
MoE(x) = \sum_{i=1}^{k}g_i(x)\cdot f_i(x),
\end{equation}
where $f_i(x)$ indicates the output produced by the expert $i$. The gating function can be implemented as a random selector or neural network. 

\begin{table}[t]
\centering
\footnotesize
\setlength{\tabcolsep}{20pt}{
\begin{tabular}{l c c}
\toprule
\textbf{Dataset} & \textbf{AdamW} & \textbf{OMoE}\\
\midrule
CoLA & 3.01E-06 & 3.07E-06 \\

SST-2 & 1.63E-05 & 1.74E-05 \\

MRPC & 5.12E-05 & 6.52E-05 \\

STS-B & 1.76E-06 & 1.92E-06 \\

QQP & 8.98E-05 & 3.11E-04 \\

MNLI & 9.03E-05	& 2.08E-04 \\

QNLI & 2.38E-05	& 2.95E-05 \\

RTE & 7.23E-05 & 1.67E-04 \\
\bottomrule
\end{tabular}}
\caption{
 The variance of the parameters over experts in BERT-MoE.
}
\label{tab:similarity1}
\end{table}

\subsection{Orthogonal Weights Modification (OWM)}
OWM (Orthogonal Weight Modification)~\cite{zeng2019continual} is a learning algorithm that aims to address the issue of catastrophic forgetting in continual learning. Catastrophic forgetting occurs when a model trained on new tasks interferes with previously learned knowledge, resulting in a significant decline in the performance of the previously learned tasks. OWM enhances the model's generalization ability by capturing task-specific features while preserving previously learned knowledge. A crucial step in protecting existing knowledge during network training is to create an appropriate orthogonal projection that operates in a space perpendicular to the one defined by previous inputs. This enables the network to shield previously acquired knowledge effectively. OWM constructs an orthogonal projector~\cite{ben2003generalized,yanai2011projection,haykin2008adaptive} as $A(A^TA)^{-1}A^T$, where $A$ contains previous inputs as columns. To avoid the matrix invertibility problem, OWM adds a small constant $\alpha$. So the projector will be $A(A^TA+\alpha I)^{-1}A^T$. However, as $A$ includes all previously trained input vectors, to update $P$, we need to recalculate it after incorporating the new input into $A$. Therefore, OWM transform the projector in OWM to RLS projector~\cite{haykin2008adaptive}.

Specifically, we consider a feed-forward network of $L+1$ layers. The orthogonal projector of layer $l$ is initialized as $\textbf{I}_l$, where $\textbf{I}_l$ refers to a unit matrix and will be updated when one task is finished. For the $i^{th}$ batch during training the $j^{th}$ task, the orthogonal projector of layer $l$ after training $j-1$ tasks is represented as:
\begin{equation}
\begin{aligned}
    &\mathbf{P}_l(i,j) =\mathbf{P}_l(i-1,j) \\
    &-\mathbf{k}_l(i,j) \overline{\mathbf{x}}_{l-1}(i,j)^T \mathbf{P}_l(i-1,j)\\   
\end{aligned}
\label{eq:OWMP1}
\end{equation}

\begin{equation}
\begin{aligned}
&\mathbf{k}_l(i,j) =\mathbf{P}_l(i-1,j) \overline{\mathbf{x}}_{l-1}(i,j) / \\ 
&\left[\alpha + \overline{\mathbf{x}}_{l-1}(i,j)^T \mathbf{P}_l(i-1,j) \overline{\mathbf{x}}_{l-1}(i,j)\right],
\end{aligned}
\label{eq:OWMP2}
\end{equation}

where $\mathbf{x}_{l-1}$ is the output of the ${l-1}^{th}$ layer in response to the mean of the inputs in the $i^{th}$ batch of $j^{th}$ task. $\alpha$ refers to decaying factor as $\alpha(i,j) = \alpha_0\lambda^{i/j}$ for the $i^{th}$ batch of data in the $j^{th}$ task.

The orthogonal projector pushes the parameters update to the orthogonal direction of the previous tasks. By using gradient descent to find a suitable weight configuration, the OWM helps the network to learn new tasks without compromising the performance of tasks it has already learned. Combined with SGD, the parameter update will be:

\begin{equation}
\begin{aligned}
\mathbf{W}_l(i, j) &=\mathbf{W}_l(i-1, j)\\
&+\kappa(i, j) \Delta \mathbf{W}_l^{B P}(i, j) \quad \text { if } \quad j=1 
\end{aligned}
\label{eq:OWM1}
\end{equation}

\begin{equation}
\begin{aligned}
\mathbf{W}_l(i, j)&=\mathbf{W}_l(i-1, j)\\
&+\kappa(i, j) \mathbf{P}_l(j-1) \Delta \mathbf{W}_l^{B P}(i, j) \quad \\
&if \quad j=2,3, \cdots,
\end{aligned}
\label{eq:OWM2}
\end{equation}
where $\mathbf{W}_l(i,j)$ is the $l^{th}$ layer after training by $j-1$ tasks and $i$ batches in $j^{th}$ task, $\kappa(i, j)$ is the learning rate, $\Delta \mathbf{W}_l^{B P}(i, j)$ is the gradient calculated by $i^{th}$ batch in the $j^{th}$ task. In general, the procedure of OWM will be:

\begin{enumerate}
  \item Initialization of parameters: initialize $\mathbf{W}_l(0)$ and $\mathbf{P}_l(0)$
  \item Forward propagate the inputs of $i^{th}$ batch in the $j^{th}$ task. then back propagate the errors and calculate weight modification $\Delta \mathbf{W}_l^{B P}(i, j)$
  \item Update the weight matrix in each layer by Equation~\ref{eq:OWM1} and Equation~\ref{eq:OWM2}.
  \item Repeat steps 2) to 3) for the next batch.
  \item When $j^{th}$ task is finished, forward propagate the mean of the inputs for each batch in the $j^{th}$ task successively. Then update $\mathbf{P}_l$ by Equation~\ref{eq:OWMP1} and Equation~\ref{eq:OWMP2}.
  \item Repeat steps 2) to 5) for the next task.
\end{enumerate}

\section{Orthogonal Optimizer for MoE}

\subsection{Degeneracy in MoE}
Gao et al.\cite{DBLP:conf/coling/Gao0ZLW22} and Shen et al.\cite{shen2019mixture} have highlighted several issues in MoE architecture, including the degeneracy caused by \textit{imbalance} problem and \textit{homogenous representation} problem. While the \textit{imbalance} problem has been explored in previous research, we specifically focus on the \textit{homogenous representation} problem in this study. Our experiments also confirm the existence of this issue, as illustrated in Table~\ref{tab:similarity1}. Table~\ref{tab:similarity1} displays the similarity between the experts in the MoE architecture after fine-tuning the BERT model on the GLUE dataset. The weights of the experts are directly copied from the pre-trained model. The small variation in the similarity scores indicates that the experts in the MoE architecture have not learned diverse knowledge that is unique to specific inputs. 
We consider parameters to be \textit{similar} if their differences at the same position in different experts are below a certain threshold ($threshold = 1E-3$). Our analysis reveals that the percentage of similar parameters in the fine-tuned model exceeds 99\%. Consequently, this often leads to MoE models underperforming compared to the original model.

\begin{figure*}[t]
\centering
\includegraphics[width=0.96\linewidth]{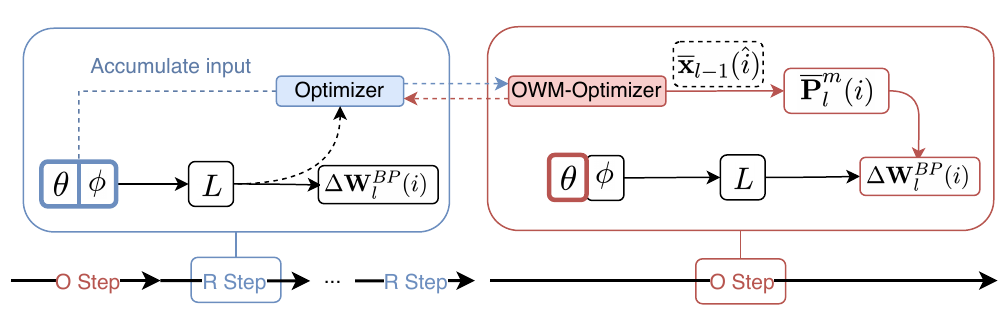}
\caption{\textbf{The full training process of OMoE}. OMoE consists of two optimizers: the base optimizer (the blue \textit{Optimizer} in the figure) and the OWM optimizer (the red \textit{OWM-Optimizer} in the figure). The training process also consists of 2 kinds of alternative steps: \textbf{R Step} (correspondents to the accumulating phase) and \textbf{O Step} (correspondents to the orthogonal phase). In \textbf{R Step}, $\Delta \mathbf{W}_l^{BP}(i)$ directly guides the update of parameters $\theta$ and $\phi$. In \textbf{O Step}, average orthogonal projector $\mathbf{\overline{P}}^{m}_l(i)$ is calculated based on average input $\overline{\mathbf{x}}_{l-1}(i)$ and then guide the gradient to be orthogonal. The two optimizers share the states like momentum and hyperparameters.}
\label{fig:trainprocess}
\end{figure*}

\subsection{OWM for Experts}

Inspired by Orthogonal Weight Modification (OWM) in multitask learning, we adapt it to MoE models. We introduce OMoE, an expert optimizer for MoE. To align with the concept of \textit{tasks} in multitask learning, we divide the training process into two distinct phases: the \textit{accumulating phase} and the \textit{orthogonal phase}.

During the accumulating phase, the model undergoes regular updates, while the optimizer simultaneously accumulates representations that capture the subspace associated with each expert. These representations are learned by observing the inputs of all experts within the same mini-batch. In essence, the optimizer computes the input from all experts and utilizes it to represent the subspace specific to each expert. During the orthogonal phase, the parameters of each expert are updated in a direction that is orthogonal to the subspace defined by the previously learned inputs of other experts. This guarantees that each expert captures distinct and non-overlapping facets of the data. To prevent updates in an orthogonal direction unrelated to the representative subspace, experts first acquire initial knowledge in the accumulating phase before entering the orthogonal phase.

To summarize, our OMoE employs an alternating training strategy with different optimizers, which allows experts to update their parameters in orthogonal directions from the subspaces of other experts, resulting in proper expert diversity and improved model performance. The accumulating phase corresponds to the \textbf{R}(egular) Step in the updating process, and the orthogonal phase corresponds to the \textbf{O}(WM) Step.

\textbf{R Step}: update all parameters by other optimizers like Adam: 
\begin{equation}
\begin{aligned}
\mathbf{W}_l(i;\theta,\phi)&=\mathbf{W}_l(i-1;\theta,\phi)\\
&+\kappa(i) \Delta \mathbf{W}_l^{B P}(i;\theta,\phi) 
\end{aligned}
\end{equation}

\ignore{
In the R Step, all parameters, $\theta$ and $\phi$, will be updated. To be more specific, parameters other than the experts (i.e., $\phi$) will be updated using the standard optimizer, while all experts (i.e., $\theta$) will accumulate input via the optimizer. The subspace spanned by these inputs will guide the update of the projector during the O step, which in turn assists with the updates to the experts.
}

\textbf{O Step:} For the $i^{th}$ update step in a accumulate phase, layer $l$ and an expert $\theta^m$ belongs to all $M$ experts, we have:
\begin{equation}
\begin{aligned}
\mathbf{W}_l(i;\theta^m)&=\mathbf{W}_l(i-1;\theta^m)\\
&+\kappa(i) \mathbf{\overline{P}}_l^m(i) \Delta \mathbf{W}_l^{B P}(i;\theta^m)
\end{aligned}
\end{equation}
\begin{equation}
\begin{aligned}
\mathbf{\overline{P}}^{m}_l(i) = \frac{1}{M}\sum_{j=1}^M{\mathbf{P}^{j}_l(i)}, j\neq m 
\end{aligned}
\end{equation}
where $\kappa(i)$ is the learning rate in $i^{th}$ update step. ${\overline{P}}^{m}_l(i)$ is calculated by the average orthogonal projector of other experts except $m$. We can get an orthogonal projector of expert $m$ by:
\begin{equation}
\begin{aligned}
\mathbf{P}^{m}_l(i) &=\mathbf{P}^{m}_l(i-1)\\
&-\mathbf{k}^{m}_l(i) \overline{\mathbf{x}}_{l-1}(\hat{i})^T \mathbf{P}^{m}_l(i-1)
\end{aligned}
\end{equation}

\begin{equation}
\begin{aligned}
\mathbf{k}^{m}_l(i) &=\mathbf{P}^{m}_l(i-1) \overline{\mathbf{x}}_{l-1}(\hat{i}) / \\
&\left[\alpha + \overline{\mathbf{x}}_{l-1}(\hat{i})^T \mathbf{P}^{m}_l(i-1) \overline{\mathbf{x}}_{l-1}(\hat{i})\right],
\end{aligned}
\end{equation}
where $\alpha$ decaying as $\alpha_0 \lambda^{i /  {n}}$ for $i$th mini-batch in all $n$ batches. Noted that $i-s\leq\hat{i}\leq i$. The project must be updated for \textbf{all} accumulating phase. 
In O Step, only $\theta$ will be updated.\\

Hidden states are shared between the ordinary optimizer and our OMoE optimizer. We define skipping step $s$ in Algorithm~\ref{algo:trainprocess}. The algorithm makes O updates every $s$ epoch while always taking R steps. The capacity of the projector limits the skipping step to being too small. Let us consider a task that has $U$ update steps. The capacity of the projector can be viewed as the \textit{rank} of $\mathbf{P}_i$, where i is the $0\leq i\leq U/s$. In O Step, we update $\mathbf{P}_i$ by $\mathbf{P}_{i+1} = \mathbf{P}_{i} -\Delta\mathbf{P}_{i+1} $. In ideal condition that every phase is independent, $\textit{range}(\mathbf{P}_i) \bigcap \textit{range}(\Delta\mathbf{P}_i) = \emptyset$. Therefore $\textit{rank}(\mathbf{P}_{i+1}) = \textit{rank}(\mathbf{P}_{i}) - \textit{rank}(\Delta\mathbf{P}_{i+1})$. It is obvious that with $i$ increase, the $\mathbf{P}_{i}$ becomes close to 0 and if $\mathbf{P}_{i}$ is 0, the projector will not be able to represent the subspace. So $s$ can not be set too small. In experiments, we set $s$ as 5.

\begin{algorithm}[ht]
    \caption{The process of \textbf{O Step} for OMoE, with with optimizer $\mathcal{O}$}
    \begin{algorithmic}[1] 
    \Procedure{OWM-O}{$\textbf{X}$;$\mathcal{O}$}
    \State $Z_{MoE} = \sum_{i=1}^{k}g_i(x)\cdot f_i(x;\theta)$
    \State $Z = F(Z_{MoE};\phi)$
    \State Forward propagation with $Z$ and calculate $L$.
    \State Calculate the gradient $\Delta \mathbf{W}_l^{B P}(i;\theta,\phi)$  with back propagation for layer $i$.
    \State Update the projector by all accumulated inputs: $\mathbf{P}^{m}_l(i) =\mathbf{P}_l(i-1)-\mathbf{k}^{m}_l(i) \overline{\mathbf{x}}_{l-1}(\hat{i})^T \mathbf{P}^{m}_l(i-1)$
    \State Calculate the projector by averaging the projector of other experts: $\mathbf{\overline{P}}^{m}_l(i) = \frac{1}{M}\sum_{j=1}^M{\mathbf{P}^{j}_l(i)}, j\neq m$
    \State $\theta \leftarrow \theta - \eta \cdot \mathbf{\overline{P}}^{m}_l(i) \cdot \Delta \mathbf{W}_l^{B P}(i;\theta)$ \Comment{Update experts parameters}
    \EndProcedure
    \end{algorithmic}
\label{algo:g_step}
\end{algorithm}

\begin{algorithm}[ht]
    \caption{The process of \textbf{R Step} for OMoE, with optimizer $\mathcal{R}$}
    \begin{algorithmic}[1] 
    \Procedure{OWM-R}{$\textbf{X}$;$\mathcal{R}$}
    \State $Z_{MoE} = \sum_{i=1}^{k}g_i(x)\cdot f_i(x;\theta)$
    \State $Z = F(Z_{MoE};\phi)$
    \State Forward propagation with $Z$ and calculate $L$.
    \State Calculate the gradient $\Delta \mathbf{W}_l^{B P}(i;\theta,\phi)$  with back propagation for layer $l$.
    \State $\theta \leftarrow \theta - \eta \cdot \Delta \mathbf{W}_l^{B P}(i;\theta,\phi)$
    \State $\phi \leftarrow \phi - \eta \cdot \Delta \mathbf{W}_l^{B P}(i;\theta,\phi)$ \Comment{Update all parameters}
    \EndProcedure
    \end{algorithmic}
\label{algo:r_step}
\end{algorithm}

\begin{algorithm}[ht]
    \caption{Training process of OMoE, at mini-batch $e$, with data $\mathcal{D}$, skipping step $s$, OMoE optimizer $\mathcal{O}$ and regular optimizer $\mathcal{R}$}
    \begin{algorithmic}[1] 
    \Procedure{Opti}{$\textbf{X}$, $s$, $e$, $\mathcal{O}$, $\mathcal{R}$}
    \For{\textbf{X} in $\mathcal{D}$}
    \If{$e$ mod $s$ = 0}
    \State OMoE-O($\textbf{X}$) \Comment{Do orthogonal update}
    \Else
    \State OMoE-R($\textbf{X}$)
    \State $\mathcal{O}$.add($\textbf{X}$) \Comment{Accumulate Input in OMoE optimizer}
    \EndIf 
    \EndFor
    \EndProcedure
    \end{algorithmic}
\label{algo:trainprocess}
\end{algorithm}

\section{Experiment}
In this section, we empirically evaluate OMoE by fine-tuning the pre-trained language model on various benchmarks and tasks. 
Firstly, we introduce the compared base models.
Then, we evaluate our proposed method in the GLUE benchmark ~\cite{DBLP:conf/iclr/WangSMHLB19}, the SuperGLUE benchmark~\cite{wang2019superglue}, the QA task~\cite{2016arXiv160605250R}, and the NER task~\cite{sang2003introduction}. 

\subsection{Compared Models}
The proposed OMoE is evaluated on three language models:
\begin{itemize}
\item \textbf{BERT}~\cite{devlin-etal-2019-bert} is a transformer-based pre-trained model consisting of 12 encoders, each with 12 bidirectional self-attention heads (BERT-BASE). The MoE version of the BERT base model composes 4 experts per layer.
\item \textbf{RoBERTa}~\cite{liu2019roberta} builds on BERT’s language masking strategy and introduces key modifications, the removal of the next-sentence prediction objective, the use of larger mini-batches and higher learning rates, as well as longer training on a larger dataset.
\item \textbf{ALBERT}~\cite{lan2019albert}, which is known as \textit{A Lite BERT}, employs a number of innovative techniques, such as factorized embedding parameterization, cross-layer parameter sharing, and parameter-sharing across layers, to reduce the number of parameters required by the model and improve its efficiency, while maintaining or even surpassing the performance of BERT.
\end{itemize}

Specifically, to evaluate the effectiveness of OMoE, we conduct experiments on three popular transformer-based models: BERT-base, RoBERTa-base, and ALBERT-base-v2. To convert these models into MoE models, we replace all layers with MoE layers, each consisting of four experts. To initialize the experts, we follow the common practice in previous works \cite{gupta2022sparsely, ye2022eliciting,DBLP:conf/coling/Gao0ZLW22} by replicating the pre-trained weights of the feed-forward network.

Our evaluation focuses on the fine-tuning performance of the models on several benchmark datasets, including GLUE, SQuAD, SuperGLUE, and CoNLL-2003. We train the model on each dataset with a specific set of hyperparameters and evaluate its performance on the test set. We compare the results of our proposed OMoE method with the standard MoE. The fine-tuning is performed using 8 Nvidia A100 GPUs and the HuggingFace Transformers library.

\begin{table*}[t]
\resizebox{\linewidth}{!}{
\begin{tabular}{lccccccccl}
\toprule
& \bf MNLI & \bf QNLI & \bf QQP & \bf RTE & \bf SST-2 & \bf MRPC & \bf CoLA & \bf STS-B & \bf Avg \\
\midrule 
\multicolumn{10}{l}{\textit{MoE for $BERT_{Base}$}}\\
BERT-MoE$_{AdamW}$ & 83.99 & \textbf{90.54} & 90.62 & 75.81 & 90.83 & 87.22 & 53.39 & 89.05 & 82.68\\
BERT-MoE$_{OMoE}$ & \textbf{84.31} & 90.46 & \textbf{90.98} & \textbf{77.62} & \textbf{91.51} & \textbf{87.47} & \textbf{54.69} & \textbf{89.39} & \textbf{83.30}$^{\Uparrow+0.62}$ \\
\midrule
\multicolumn{10}{l}{\textit{MoE for $RoBERTa_{Base}$}} \\
RoBERTa-MoE$_{AdamW}$ & 84.72 & 90.58 & 90.91 & 74.73 & 91.62 & \textbf{88.12} & 53.39 & 89.66 & 82.96\\
RoBERTa-MoE$_{OMoE}$ & \textbf{85.42} & \textbf{90.70} & \textbf{91.36} & \textbf{79.03} & \textbf{92.64} & 87.82 & \textbf{53.9} & \textbf{90.04} & \textbf{83.86}$^{\Uparrow+0.9}$ \\
\midrule
\multicolumn{10}{l}{\textit{MoE for $ALBERT_{Base}$}} \\
ALBERT-MoE$_{AdamW}$ & 84.81 & 91.37 & 88.73 & 74.25 & 90.93 & 88.01 & 54.95 & 90.32 & 82.92\\
ALBERT-MoE$_{OMoE}$ & \textbf{85.23} & \textbf{91.72} & \textbf{88.91} & \textbf{76.61} & \textbf{92.18} & \textbf{88.43} & 54.95 & \textbf{90.68} & \textbf{83.58}$^{\Uparrow+0.66}$ \\
\bottomrule
\end{tabular}}
\caption{
Results on GLUE. CoLA is evaluated using the Matthews correlation coefficient (MCC), while the other tasks use accuracy as the metric. We select 3 models for evaluation: BERT, RoBERTa and ALBERT. Layers in all three models are replaced with four-experts-MoE-layers. AdamW is used as the base optimizer.
}
\label{tab:pretrain_result}
\end{table*}

\subsection{Fine-tune on GLUE}
\textbf{Dataset.} GLUE (General Language Understanding Evaluation)~\cite{DBLP:conf/iclr/WangSMHLB19} is a collection of natural language processing tasks designed to evaluate the performance of language models. It consists of diverse tasks that cover a wide range of language understanding capabilities, including reading comprehension, classification, and natural language inference. \\

\textbf{Result.} The experimental results of our proposed OMoE optimizer on the GLUE dataset are presented in Table~\ref{tab:pretrain_result}, where we compare its performance with the AdamW optimizer. The results show that our OMoE optimizer outperforms the baseline on 7 out of 8 datasets, achieving an average score of 83.30 on BERT, 83.86 on RoBERTa, and 83.58 on ALBERT. In contrast to the AdamW optimizer, our OMoE optimizer effectively enlarges the diversity among experts, leading to an improvement of 0.62, 0.9, and 0.66 for the three models, respectively. However, we surprisingly observe degradation in performance for some low-resource tasks such as MPRC and CoLA. Upon analysis, we found that these tasks are particularly challenging for the experts in MoE to extract useful knowledge. Therefore, the OMoE optimizer cannot accurately identify the input subspace and hence cannot guarantee improvement for these tasks.

\textbf{Similarity over Experts.} 
As MoE-based Transformers are relatively new and there are limited publicly available pre-trained models, there is currently no standardized evaluation method for assessing the similarity over experts. In this study, we propose two metrics to evaluate the diversity among experts in the MoE model. Firstly, we calculate the percentage of different parameters between two experts, which provides a quantitative measure of how distinct the two experts are from each other. Specifically, we randomly select two experts and compute the ratio of the different parameters to the total number of parameters. A higher percentage indicates that the two experts are more dissimilar. Secondly, we use variance as a metric to assess the diversity among all the experts. We calculate the variance of the outputs of all experts for a given input sample. A higher variance suggests that the experts are more diverse in their predictions for the given input, which is desirable for the MoE model. These two metrics provide complementary information on the diversity among experts and can help to diagnose the degeneracy problems in MoE models.

\begin{figure}[t]
\centering
\includegraphics[width=0.8\linewidth]{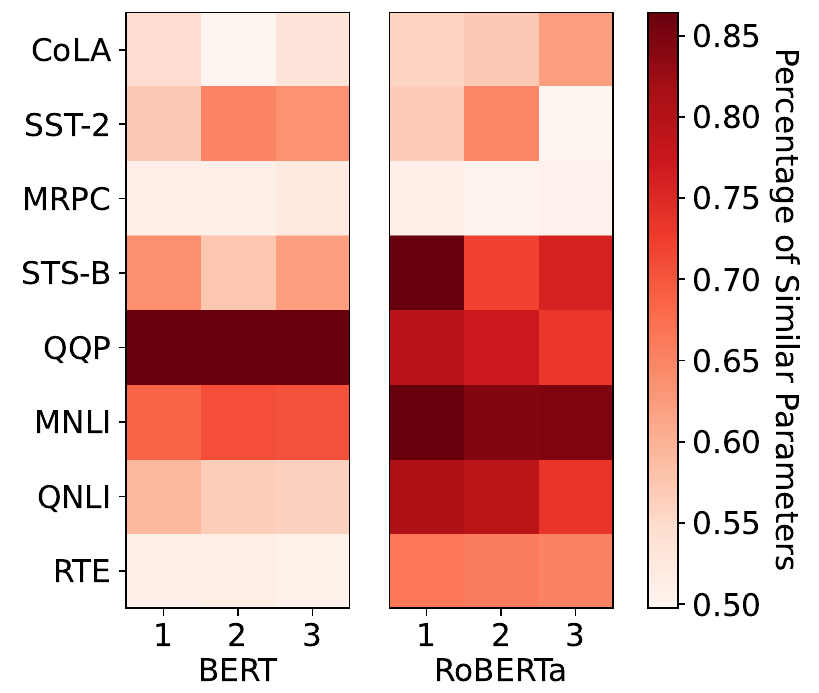}
\caption{The figure shows to what extent OMoE expands the difference between experts. The first 3 columns are for experts in BERT while the other 3 columns are for experts in RoBERTa. The depth of color in the figure represents the percentage of parameters with a larger difference in OMoE compared to AdamW.}
\label{fig:similarity}
\end{figure}

Figure.~\ref{fig:similarity} shows the percentage of different parameters. We calculated the difference over expert parameters in the model optimized by OMoE and AdamW at the same position. Based on the difference, we recorded the proportion of parameters that has a more significant difference compared with the baseline. We define the proportion as \textit{diverse degree}, indicating the gap between experts compared with the baseline. We can see that in all tasks, the diverse degree is improved compared to the baseline. Among them, the diverse degree in the QQP and MNLI tasks is more considerable, showing the powerful role of the OMoE algorithm in improving the difference for parameters. We also analyze the variance of experts. Table~\ref{tab:similarity1} shows the variance of all experts relative to the mean value. Compared to the AdamW optimizer, our optimizer effectively enhances the variance of the expert, thereby improving the performance of the MoE architecture. It is worth noting that the greater difference between experts does not necessarily mean better model performance. One possible explanation is that high-resource tasks can take advantage of a large number of parameters in a sparse model, thereby improving model performance without increasing expert differences. We will discuss the relationship between similarity and performance in Section~\ref{sec:skipstep}. \\

\subsection{Fine-tune on SuperGLUE}
\textbf{Dataset.} SuperGLUE~\cite{wang2019superglue} (General Language Understanding Evaluation) is a benchmark dataset for evaluating the performance of NLU models on a diverse set of eight language understanding tasks.

\textbf{Result.} As shown in Table~\ref{tab:super_result}. The comparison to AdamW suggests that our approach significantly improves over previous state-of-the-art methods. For example, our OMoE optimizer achieves 1.7 increase in accuracy on the COPA task. Our method also yields comparable results in other experiments. 

\begin{table}[t]
\resizebox{\linewidth}{!}{
\begin{tabular}{lccccc}
\toprule
& \bf COPA & \bf BoolQ & \bf MultiRC & \bf WiC & \bf CB  \\
\midrule 
\multicolumn{6}{l}{\textit{MoE for $BERT_{Base}$}}\\
BERT-Dense & 70.0 & 71.5 & 68.4 & 65.3 & 82.1\\
BERT-MoE$_{AdamW}$ & 69.2 & 71.8 & 68.7 & 65.9 & 82.3 \\
BERT-MoE$_{OMoE}$ & \textbf{70.9} & \textbf{72.3} & \textbf{69.2} & \textbf{66.8} & \textbf{82.8} \\
\midrule
\multicolumn{6}{l}{\textit{MoE for $RoBERTa_{Base}$}} \\
RoBERTa-Dense & 82.8 & 81.8 & 72.7 & 69.5 & 88.7 \\
RoBERTa-MoE$_{AdamW}$ & 83.5 & 82.5 & 72.5 & 70.2 & 88.7\\
RoBERTa-MoE$_{OMoE}$ & \textbf{84.4} & \textbf{82.9} & \textbf{73.5} & \textbf{70.6} & \textbf{89.3}\\
\midrule
\multicolumn{6}{l}{\textit{MoE for $ALBERT_{Base}$}} \\
ALBERT-Dense & 65.3 & 72.6 & 62.6 & 61.4 & 69.4 \\
ALBERT-MoE$_{AdamW}$ & 65.7 & 72.8 & 63.0 & 61.8 & 69.1\\
ALBERT-MoE$_{OMoE}$ & \textbf{65.5} & \textbf{73.8} & \textbf{63.4} & \textbf{62.1} & \textbf{69.8}\\
\bottomrule
\end{tabular}}
\caption{
Performance on SuperGLUE of BERT, RoBERTa and ALBERT.}
\label{tab:super_result}
\end{table}

\subsection{Fine-tune on Question-Answering and Named Entity Recognition}
\textbf{Dataset}. SQuAD1.1~\cite{2016arXiv160605250R} is based on a set of over 100,000 questions from Wikipedia articles, along with their corresponding answers. SQuAD1.1 is widely used in natural language processing to train and evaluate QA models. CoNLL-2003~\cite{sang2003introduction} is a shared task for named entity recognition and multi-lingual named entity recognition. The task has become a benchmark for evaluating the performance of named entity recognition systems.\\

\begin{table}[h]
\centering
\begin{tabular}{lcc}
\hline
Model       & AdamW & OMoE \\ \hline
BERT & 87.9 & /        \\ 
RoBERTa & 89.6 & /        \\ 
ALBERT & 87.2 & /        \\ 
BERT-MoE    & 88.4 & \textbf{88.8}$^{\Uparrow+0.4}$     \\
RoBERTa-MoE & 89.3 & \textbf{89.9}$^{\Uparrow+0.6}$     \\
ALBERT-MoE & 89.3 & \textbf{88.1}$^{\Uparrow+0.9}$          \\ \hline
\end{tabular}
\caption{F1 scores result on SQuAD1.1. }
\label{tab:qa_result}
\end{table}

Table~\ref{tab:qa_result} and Table~\ref{tab:NER_result} presents the evaluation results on SQuAD1.1 and CoNLL-2003. Overall, compared to AdamW optimizer, our OMoE achieves competitive performance for QA and NER tasks.

\begin{table}[h]
\centering
\begin{tabular}{lcc}
\hline
Model       & AdamW & OMoE \\ \hline
BERT & 98.4 & /        \\ 
RoBERTa & 98.8 & /        \\ 
ALBERTA & 98.5 & /        \\ 
BERT-MoE    & 98.4 & \textbf{98.6}$^{\Uparrow+0.2}$     \\
RoBERTa-MoE & 98.5 & \textbf{98.6}$^{\Uparrow+0.1}$          \\
ALBERT-MoE & 98.5 & \textbf{98.6}$^{\Uparrow+0.1}$          \\\hline
\end{tabular}
\caption{Acc. Result on CoNLL-2003.}
\label{tab:NER_result}
\end{table}

\section{Analysis}
In this section, we provide a detailed analysis of how the skipping step and the number of experts affect the diversity and performance of models. Besides, we analyze the overhead of OMoE, including the additional storage space and the computational complexity. Finally, we discuss the importance of diversity in MoE. 

\subsection{Ablation Study: Effects of Skipping Step}
\label{sec:skipstep}

\begin{figure}[h]
\centering
\includegraphics[width=\linewidth]{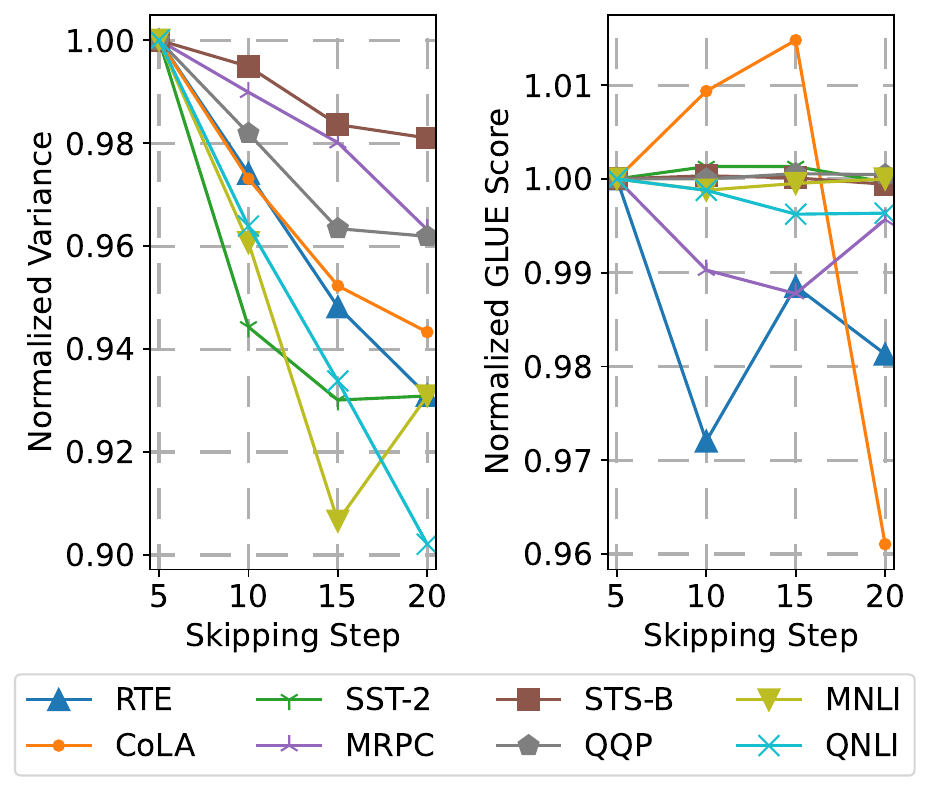}
\caption{The normalized variance of parameters in experts with different skipping steps and the normalized GLUE scores with different skipping steps.}
\label{fig:variance_and_result_without_varience}
\end{figure}

We analyze how different skipping steps affect the result, as shown in Figure~\ref{fig:variance_and_result_without_varience}. As the skipping step increases, the variance shows a trend of gradually decreasing, which also confirms our previous speculation: our method can effectively increase the difference between experts. A smaller skipping step indicates the OMoE optimizer is more frequently applied. By projecting inputs from different training phases on the projector, the projector preserves the subspace spanned by tokens processed by different experts, effectively enhancing the representation ability of each expert. However, the corresponding results show that increased expert differences do not necessarily mean better results. For example, in the CoLA task, the best results appear when skipping step = 15. Some works~\cite{DBLP:conf/coling/Gao0ZLW22} have confirmed the role of similar parameters existing among different experts. Furthermore, the additional parameters in the MoE layer possess a significant amount of information capacity and have a great impact on model performance. The results demonstrate this idea. Therefore, an alternating training strategy is necessary.

\subsection{Ablation Study: Number of Experts}

\begin{figure}[t]
\centering
\includegraphics[width=\linewidth]{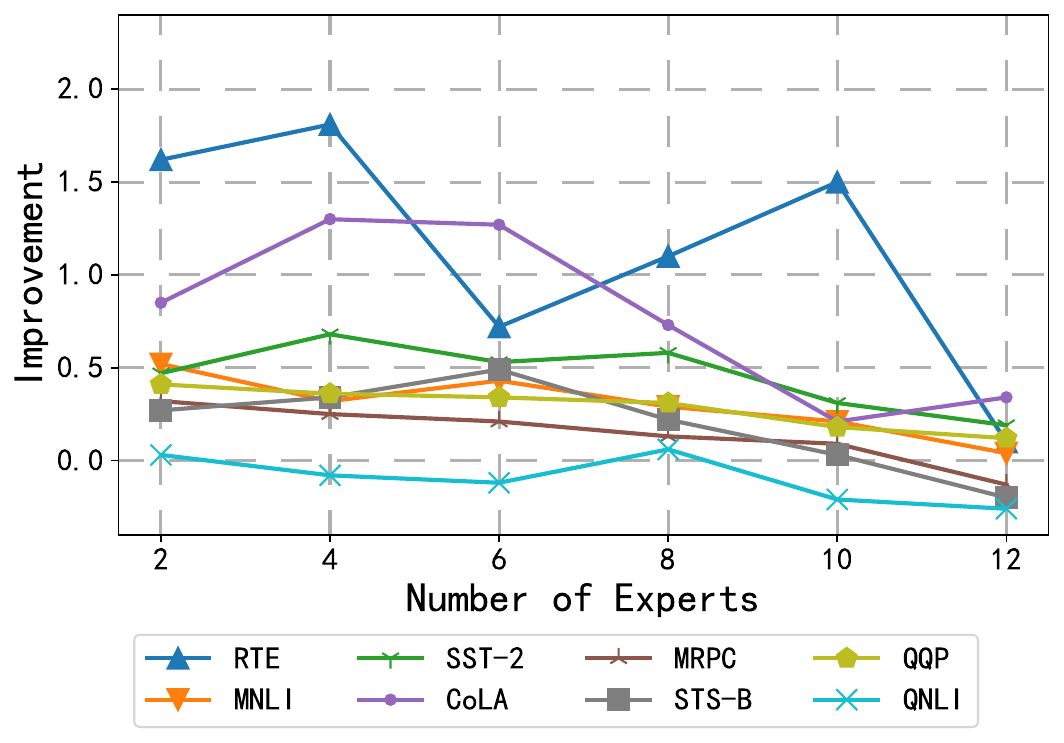}
\caption{GLUE score improvement with different numbers of experts using OMoE optimizer. In general, the benefit of orthogonal updates will become minor as the number of experts increases.}
\label{fig:num_expert}
\end{figure}

The number of experts in the MoE architecture is a crucial hyperparameter affecting the OMoE optimizer's performance. We conducted experiments on fine-tuning BERT by varying the number of experts. As shown in Figure~\ref{fig:num_expert}, the network performance remains acceptable until the number of experts is increased to a specific value, beyond which the average subspace becomes too complex to be orthogonal. This suggests that the effectiveness of OMoE optimizer may saturate beyond a certain number of experts, and adding more experts may not provide further performance improvement. Moreover, the increasing number of experts also means an increase in network capacity, which can lead to overfitting. We noticed that the input sent to each expert becomes fewer when there are more experts in the MoE architecture, as the tokens within a mini-batch are fixed. This results in insufficient input data being collected by the projector to represent the subspace accurately. Hence, the OMoE optimizer may not work well when the number of experts is too large, and the projector may fail to capture the necessary information to ensure optimal performance.

\subsection{Ablation Study: Kind of Optimizer}
Our proposed OMoE can improve the performance of various existing optimizers on MoE networks. We selected AMSGrad and Adagrad optimizers and fine-tuned BERT model to demonstrate generalizability of OMoE optimizer. The results in Table~\ref{tab:different_opti_result_MRPC} show that the optimizers modified by OMoE perform better than the unmodified ones. This indicates that our method is adaptable to different optimizers. Additionally, OMoE can leverage the latest research advances in optimizer techniques to improve performance continuously.

\begin{table}[h]
\centering
\begin{tabular}{lcc}
\hline
Optimizer       & Initial Version & OMoE Version \\ \hline
AdamW & 87.22 & \textbf{87.47}        \\ 
RMSProp & 87.15 & \textbf{87.31}         \\ 
Adagrad & 86.83 & \textbf{87.04}         \\\hline
\end{tabular}
\caption{Different optimizers result of fine-tuning BERT on MRPC.}
\label{tab:different_opti_result_MRPC}
\end{table}

\subsection{Overhead Analysis}

Our proposed OMoE optimizer requires additional storage space compared to traditional optimizers, as well as additional computing overhead during training. As shown in Table~\ref{tab:overhead}, the size of the OMoE optimizer and projector is the same as that of all experts, leading to a modest increase in storage requirements. However, this overhead is not significant (1.38$\times$) as the state of the two optimizers can be shared, and only one projector needs to be stored additionally.

In terms of computational overhead, the additional cost is 1.82GMacs (number of multiply–accumulate operations) during the training process. This overhead is acceptable compared to the high requirements of computational resources in backpropagation. The computational complexity of our method is $O(kN_{e}N^2_w)$, where $k$ is the number of experts, $N_{e}$ is the number of neurons per expert, and $N^2_w$ is the number of input weights per neuron. Overall, the additional overhead introduced by our method is affordable and does not significantly impact training efficiency.

\begin{table}[h]
\centering
\footnotesize
\begin{tabular}{cllll}
\hline
\multirow{2}{*}{Model}    & \multirow{2}{*}{Optimizer} & \multirow{2}{*}{FLOPs} & \multicolumn{2}{c}{Memory}  \\ \cline{4-5} 
                      &   &   & Model        & Optimizer \\ \hline
\multicolumn{1}{l}{T} & R & 0 & 1$\times$    & 1$\times$ \\
T-MoE                 & R & 0 & 2.55$\times$ & 1$\times$ \\ 
\multicolumn{1}{l}{T-MoE} & R+O                        & 1.82GMacs              & 2.55$\times$ & 1.38$\times$ \\ \hline
\end{tabular}
\caption{The overhead of our proposed method. T refers to the Transformer model. \textit{R} refers to updating the model using \textbf{R Step} while \textit{O} refers to updating the model using \textbf{O Step}. The \textit{FLOPs} refers to additional FLOPs introduced by our method.}
\label{tab:overhead}
\end{table}

\subsection{Discussion: Diversity and Performance}
In the Motivation section, we pointed out that the unsatisfactory performance of MoE is due to the excessive similarity over parameters of experts, which contradicts the original intention of MoE. However, it does not necessarily mean that the larger the difference between the parameters of the experts, the better the performance of the MoE. 

Firstly, although the Gating Function assigns different tokens to different experts, these tokens in the same dataset should follow the same distribution. Therefore, the parameters of experts should be similar to a certain extent. Moreover, overly different parameters can lead to instability and inconsistency in the behavior of the MoE, which can further lead to poor performance. 

Secondly, in the OMoE algorithm, the \textbf{O Step} and \textbf{R Step} run alternately, and experts do not always make orthogonal updates. This ensures that the differences between experts should not expand infinitely, and prevents the parameters of the experts from becoming too different. Furthermore, research on the Gating Function has shown that the correctness of the router has a significant impact on performance. If we make the parameters fully orthogonal based on the possibly inaccurate routing result, it may violate the original intention of MoE and lead to poor performance.

Overall, increasing the difference between the parameters of the experts should not be the primary goal of MoE. Instead, the focus should be on finding the optimal balance between the similarity and differences of the parameters to achieve the best performance. The experiments have also confirmed this view, showing that excessively different parameters can lead to performance degradation.

\section{Related Work}

\paragraph{PLMs with MoE}
MoE~\cite{jacobs1991adaptive} was first proposed in 1991 and has been widely applied in various fields~\cite{dai-etal-2022-stablemoe,zhou2022mixture,he2023mera}. MoE Transformers represent a novel approach to enhancing the performance of transformers. Unlike traditional feed forward blocks, Sparse MoE blocks consist of two key components: a gating function~\cite{DBLP:conf/iclr/ShazeerMMDLHD17} and a collection of feed forward neural network experts~\cite{riquelme2021scaling}. The gating function serves as the primary control mechanism for assigning tokens to specific experts. It generates a sparse output that allocates each token to a specific expert. This allocation is based on the token's properties and the expertise of the available experts. As a result, each expert specializes in a particular type of token, which allows them to process those tokens more effectively. After the advent of pre-trained language models~\cite{devlin-etal-2019-bert,liu2019roberta,DBLP:conf/iclr/ClarkLLM20, DBLP:conf/iclr/LanCGGSS20,Zhong2022TowardEL,zhong2022e2s2}, the \textit{Gating-Expert} architecture~\cite{fedus2021switch,du2022glam,DBLP:conf/iclr/LepikhinLXCFHKS21,lewis2021base} was quickly applied to PLMs and has achieved many successes. These studies improve training efficiency of PLMs with MoE by designing new routing strategies and introducing more parameters. Our method is based on the PLMs with MoE and improves the performance by enlarging the diversity of experts.\\

\paragraph{Neural Network Optimizer} Optimizer is a method used to adjust the parameters of a neural network to minimize the loss function. Different optimizers have a significant impact on the performance of a neural network. Optimizers based on the stochastic gradient descent method are currently mainstream optimizers. For example, the Adam~\cite{DBLP:journals/corr/KingmaB14} that uses moving averages of the parameters to provide a running estimate of the second raw moments of the gradients and RMSProp that divides the learning rate for weight by a running average of the magnitudes of the recent gradients for that weight. Some optimizers like AdamW~\cite{DBLP:conf/iclr/LoshchilovH19} use regularization that adds a penalty to the loss function to prevent overfitting. Also, other sharpness-aware optimizers~\cite{Zhong2022ImprovingSM,Sun2023DynamicRS} enhance the model generalization by adding a perturbation to parameters. Recently, some optimizers~\cite{zhang2022bort} use the orthogonal constraint to improve performance. To the best of our knowledge, we are the first to introduce OWM to MoE.\\

\paragraph{Optimization for MoE} A more accurate gating mechanism is the key point to optimize MoE. Gating which is learned via backpropagation~\cite{DBLP:conf/iclr/ShazeerMMDLHD17,DBLP:conf/iclr/LepikhinLXCFHKS21}, perhaps with a regularizer to encourage load balancing across experts, is the mainstream. Roller\cite{roller2021hash} proposes a novel hash function-based gating layer, which requires no routing parameters or extra terms in the objective function such as a load balancing loss. Zhou~\cite{zhou2022mixture} proposes a method that makes experts select the top-k tokens instead of letting tokens select the top-k experts. As a result, each token can be routed to a variable number of experts and each expert can have a fixed bucket size. Recent studies~\cite{rajbhandari2022deepspeed} apply knowledge distillation for faster inference. These kinds of studies reduce the overall serving cost for inference. We propose a novel optimizer for MoE to solve degeneracy instead of the conventional methods mentioned above. We encourage the orthogonal updates for experts to enlarge the diversity.

\section{Conclusion}
In this paper, we propose an orthogonal optimizer (OMoE) for MoE-based language models to address the issue of performance degeneracy and high expert representation similarities in MoE. Specifically, we employ an alternating training strategy consisting of different phases. We enable experts to update their parameters in orthogonal directions with respect to the subspaces defined by other experts. To evaluate the effectiveness of OMoE, we conduct extensive experiments on various benchmarks, including the GLUE benchmark, the SuperGLUE benchmark, the QA task, and the NER task. The experimental results demonstrate the significant performance improvements achieved by OMoE compared to baseline methods. Notably, our work represents the first application of OWM for optimizing experts in MoE models with enhanced and appropriate diversity.



\bibliography{ecai}

\end{document}